\documentclass{article}
 \usepackage[preprint]{neurips_2026}

\usepackage[utf8]{inputenc} 
\usepackage[T1]{fontenc}    
\usepackage{hyperref}       
\usepackage{url}            
\usepackage{booktabs}       
\usepackage{amsfonts}       
\usepackage{nicefrac}       
\usepackage{microtype}      
\usepackage{xcolor}         

\usepackage{xspace}
\usepackage{amsmath}
\usepackage{colortbl}
\usepackage{graphicx}
\usepackage{multirow}
\definecolor{ourblue}{RGB}{214, 230, 255}
\definecolor{cyan}{HTML}{00D4FF}
\definecolor{orange}{HTML}{FF6B35}
\usepackage{tikz}
\usepackage{gensymb}

\newcommand{\datasetname}{{AmbientEye}\xspace}

\title{\datasetname: A Dataset for Pupil Segmentation under Natural Ambient Infrared Illumination}

\author{
  Mingyu Han\textsuperscript{1},
  Hyunyoung Han\textsuperscript{1},
  Nitheekulawatn Thommakoon\textsuperscript{1},
  Gangtae Park\textsuperscript{1},\\
  \textbf{Jieun Han\textsuperscript{1},}
  \textbf{Xucong Zhang\textsuperscript{2}}
  \textbf{and Ian Oakley\textsuperscript{1}}\\
  \textsuperscript{1}Electrical Engineering, Korea Advanced Institute of Science and Technology, KR\\
  \textsuperscript{2}Intelligent Systems Department, Delft University of Technology, NL\\
  \texttt{\{mghan, hyhan, thommakoon, rkdxo0417, ktp20, ianoakley\}@kaist.ac.kr}\\
  \texttt{xucong.zhang@tudelft.nl}\\
}

\begin{document}

\maketitle

\begin{abstract}
Eye tracking is essential for smart glasses, as it provides insight into user attention for ambient intelligence applications. However, most existing eye-tracking systems rely on active infrared (IR) illumination, creating practical barriers to all-day outdoor use due to power consumption. In this paper, we investigate whether passive IR cameras alone, without any active IR light source, can enable reliable pupil detection in unconstrained outdoor environments, where ambient sunlight serves as the sole illumination source. To support this investigation, we introduce AmbientEye, a large-scale dataset of 2,606,225 eye images collected from 35 participants from 19 countries. It is captured outdoors under natural sunlight with two off-axis camera configurations and two sun-orientation conditions. 
We provide high-quality pupil annotation through SAM2 automatic segmentation, followed by refinement by human annotators. We benchmark a state-of-the-art pupil segmentation algorithm on our dataset and compare its performance with that on existing datasets under controlled IR illumination. Results reveal a substantial drop in pupil segmentation performance from 0.928 on controlled IR datasets to 0.767 on AmbientEye. This performance gap highlights the challenge of the ambient-light setting. This positions AmbientEye as a first benchmark for an unexplored and highly practical eye-tracking scenario.
\end{abstract}

\section{Introduction}
Smart glasses are emerging as a promising wearable platform capable of seamlessly augmenting everyday perception and interaction~\cite{grauman2022ego4d}. By aligning with the user’s visual perspective, they enable continuous and natural capture of the surrounding environment and immersive interactions, forming the foundation for ambient intelligence~\cite{aarts2006true}. 
For interaction on smart glasses, human eye gaze serves as a critical signal, providing an immediate indication of user attention~\cite{blattgerste2018advantages, plopski2022eye}. Consequently, head-mounted eye trackers integrated into smart glasses have become indispensable tools for recording oculomotor metrics such as pupil diameter and fixation duration~\cite{drews2024strategies}, with applications ranging from user behavior analysis~\cite{2023DualMobilemayrand} and LLM-based assistance~\cite{konrad2024gazegpt} to cognitive state measurement~\cite{2021RethinkingEyeblinkchob}.
Accordingly, eye-tracking technology has evolved from stationary laboratory systems to portable, glasses-based devices, including the Tobii Glasses X~\cite{TobiiGlassesX}, Pupil Neon~\cite{PupilNeon}, and Meta Aria~\cite{kong2025ariagen2pilot}.

However, deploying eye-tracking systems on everyday smart glasses introduces significant practical constraints. Existing approaches largely rely on active infrared (IR) illumination to form controlled lighting conditions for robust pupil detection. Therefore, most existing datasets for head-mounted eye tracking assume a well-lit eye illuminated by ample infrared light sources~\cite{santini2018pure, fuhl2016pupil, santini2018purest}. However, in real-world settings, particularly outdoors, ambient sunlight introduces strong and uncontrolled IR components that can degrade pupil contrast and violate these assumptions~\cite{rusnak2025enhancing}. 

In addition, continuous active IR illumination leads to a significant power demand, which is problematic for commercial smart glasses powered by miniature batteries. Commercial wearable eye trackers usually employ multiple IR lamps. Recent measurements indicate the IR irradiance of eye trackers ranges from 279 to 1800\,$\mu$W/cm$^2$~\cite{carminati2025energy}. To estimate the actual energy cost of this illumination, we consider a representative commercial IR LED (e.g., OSRAM SFH 4050~\cite{SFH4050}). Based on its specifications, generating this level of irradiance translates to an estimated power consumption of $4.5\,\text{mW}$ to $30\,\text{mW}$ per single LED. Given that wearable systems typically require an array of such LEDs, combined with an active IR camera (e.g., OmniVision OV6211~\cite{OV6211}, consuming $85\,\text{mW}$ at $120\,\text{fps}$), the total continuous power draw becomes prohibitively high for the strict energy budget of all-day smart glasses.

A variety of strategies were explored to reduce reliance on IR illumination in wearable eye tracking. Early efforts focused on algorithmic optimization, employing neural networks and sparse pixel sampling to lower computational demands while retaining visible-light cameras~\cite{2014ItStartszhang, 2014IShadowDesignmayberrya}. Subsequent approaches introduced environmental adaptation, enabling systems to dynamically adjust their sensing modality based on eye movement patterns and ambient lighting~\cite{2015CIDEREnablingmayberry}. 
More radical approaches eliminate cameras altogether, using photodiode arrays~\cite{2018BatteryFreeEyelia}, ultra-low-resolution sensors~\cite{2017InvisibleEyeMobiletonsen}, or alternative modalities such as acoustic sensing~\cite{2024GazeTrakExploringlia} and electrooculography~\cite{2009WearableEOGbulling} for gaze estimation. While these approaches are promising, none has yet achieved the maturity required for adoption in commercial devices. Critically, most of these approaches still aim to actively control the illumination, whether through visible light, low-power IR, or alternative modalities, leaving the regime of fully passive sensing under ambient outdoor light unexplored.

In this paper, we explore a simple yet effective approach to enable efficient eye tracking for smart glasses with ambient sunlight. Specifically, we eliminate active IR illumination and investigate the feasibility of using only IR cameras alone for eye tracking. Although indoor environments often lack dedicated IR light sources, natural sunlight in outdoor settings provides sufficient IR illumination for reliable sensing. This enables a ready-to-deploy solution for existing smart glasses platforms, where the eye-tracking system can adaptively switch IR illumination on or off depending on ambient lighting conditions.
Unfortunately, most existing datasets are predominantly collected in controlled indoor environments with active IR illumination, leaving outdoor conditions with uncontrolled illumination inherently unrepresented. In outdoor settings, ambient near-infrared irradiance from sunlight varies substantially with sun angle, cloud cover, and the user's orientation, creating a highly dynamic illumination regime that has not been fully studied. 

To bridge this gap, we present \textbf{\datasetname}, a large-scale dataset of 2,606,225 eye images collected exclusively outdoors under natural ambient sunlight illumination, without any active IR light source. Data were collected from 35 participants across 19 countries under two distinct sun-orientation conditions (facing towards and away from the sun), with two off-axis camera configurations simultaneously capturing the eye. All images were initially annotated using SAM2~\cite{ravi2024sam} segmentation model and verified by human annotators. We further benchmark a state-of-the-art pupil segmentation method on \datasetname, evaluating its robustness under these unconstrained outdoor conditions. Our experimental results show the significant challenges for eye tracking systems on smart glasses in outdoor settings without active infrared illumination. It sheds light on this promising research direction for developing low-power eye tracking methods under ambient sunlight.

\section{Related Work}

\begin{table*}[t]
\centering
\caption{A Systematic Comparison of existing near-eye pupil detection datasets.
\textbf{Camera Axis}: \emph{Off} = oblique head-mounted view; \emph{On} = frontal VR/AR HMD view.
\textbf{Light Source}: primary illumination used during capture.
\textbf{Outdoor}: includes outdoor or uncontrolled ambient light conditions.
\textbf{Camera Views}: number of synchronized cameras capturing the same eye simultaneously.
\textbf{Multiple Viewpoints}: participants fixated on structured gaze targets covering diverse gaze directions.
$\checkmark$\,=\,yes,~~$\times$\,=\,no,~~--\,=\,not reported.}
\label{tab:dataset_comparison}
\resizebox{\textwidth}{!}{%
\begin{tabular}{lcccccccc}
\toprule
\textbf{Dataset} & \textbf{Participants} & \textbf{Images}
  & \textbf{Camera Axis}
  & \textbf{Active IR}
  & \textbf{Outdoor}
  & \textbf{Camera Views}
  & \textbf{Multiple Viewpoints} \\
\midrule
Swirski~\textit{et al.}~\cite{swirski2012}
    & 2   & 600 & Off & $\checkmark$ & $\times$     & 1 & $\checkmark$ \\
ExCuSe~\cite{fuhl2015excuse}
   & 7   & $\sim$39k & Off & $\checkmark$ & $\checkmark$ & 1 & $\times$ \\
ElSe~\cite{fuhl2016else}
  & 17  & $\sim$94k & Off & $\checkmark$ & $\checkmark$ & 1 & $\times$ \\
LPW~\cite{tonsen2016lpw}
  & 22  & 131k      & Off & $\checkmark$ & $\checkmark$ & 1 & $\checkmark$ \\
NVGaze~\cite{kim2019nvgaze}
  & 35  & 2.5M      & On  & $\checkmark$ & $\times$     & 1 & $\checkmark$ \\
OpenEDS~\cite{garbin2019openeds}
  & 152 & 357k      & On  & $\checkmark$ & $\times$     & 1 & $\checkmark$ \\
MagicEyes~\cite{wu2020magiceyes}
  & 587 & 800k+     & Off & $\checkmark$ & $\times$     & 1 & $\checkmark$ \\
TEyeD~\cite{fuhl2021teyed}
  & 39+ & 20M+      & Off/On & $\checkmark$ & $\checkmark$ & 1 & $\checkmark$ \\
\midrule
\rowcolor{ourblue}
\textbf{AmbientEye (Ours)}
  & \textbf{35}  & \textbf{2.5M}
  & \textbf{Off}
  & $\boldsymbol{\times}$
  & $\boldsymbol{\checkmark}$
  & \textbf{2}
  & $\boldsymbol{\checkmark}$ \\
\bottomrule
\end{tabular}
}
\end{table*}

\subsection{Pupil Segmentation}
\label{sec:related_methods}
Pupil segmentation is the foundation of both classical and modern gaze estimation pipelines. Early methods such as ExCuSe~\cite{fuhl2015excuse} and ElSe~\cite{fuhl2016else} relied on edge filtering and ellipse fitting under the assumption of consistent IR contrast, but degraded sharply on real-world images with reflections, occlusions, or varying illumination. Learning-based methods subsequently improved robustness: PupilNet~\cite{fuhl2017pupilnet} trained convolutional networks for pupil center detection, DeepVOG~\cite{yiu2019deepvog} introduced U-Net-based pupil segmentation coupled with a 3D eyeball model, and RITnet~\cite{chaudhary2019ritnet} extended segmentation to multiple eye parts (sclera, iris, pupil). EllSeg~\cite{kothari2021ellseg} predicts ellipse parameters directly and reports substantial improvements in pupil and iris center detection over part-segmentation baselines, while EyeNet from MagicEyes~\cite{wu2020magiceyes} jointly predicts pupil center, glint, and 2D cornea center for off-axis head-mounted views. Despite this progress, all of these models are trained exclusively on datasets captured under controlled IR illumination, and their robustness under ambient outdoor IR has not been characterized, motivating a systematic benchmark under natural sunlight conditions.

\subsection{Pupil Segmentation Dataset}
A growing collection of pupil segmentation datasets has supported the development of robust pupil detection and gaze estimation methods (Table~\ref{tab:dataset_comparison}). Early off-axis datasets such as Swirski et al.~\cite{swirski2012robust}, ExCuSe~\cite{fuhl2015excuse}, and ElSe~\cite{fuhl2016else} captured head-mounted views under active IR illumination, with sample sizes ranging from hundreds to tens of thousands of images. The Labelled Pupils in the Wild (LPW) dataset~\cite{tonsen2016lpw} extended this line of work to 131k images from 22 participants under more naturalistic indoor and outdoor conditions, and remains one of the most widely used benchmarks for pupil detection in unconstrained environments. On-axis VR/AR-style datasets, including NVGaze~\cite{kim2019nvgaze} (2.5M images) and OpenEDS~\cite{garbin2019openeds} (357k images), provide large-scale, high-resolution eye images captured by frontal head-mounted cameras under tightly controlled IR lighting. MagicEyes~\cite{wu2020magiceyes} and TEyeD~\cite{fuhl2021teyed} push scale further, with TEyeD providing over 20M images across both off-axis and on-axis configurations and a wide range of annotations.

Despite this diversity, all existing datasets share a common assumption: pupil imagery is captured under active IR illumination, whether through dedicated IR LEDs co-located with the camera or through structured lab lighting. Even datasets that include outdoor recordings, such as LPW, ExCuSe, ElSe, and TEyeD, rely on active IR emitters to maintain pupil contrast against varying ambient light. Consequently, the regime in which a wearable eye tracker operates without an active IR source, relying solely on ambient sunlight as the IR illumination, is not represented in any existing benchmark. AmbientEye is, to our knowledge, the first dataset to capture this regime: 2,606,225 images from 35 participants across 19 countries, recorded outdoors under natural sunlight with two off-axis camera viewpoints and two sun-orientation conditions, providing a benchmark for studying pupil segmentation and segmentation under passive, ambient IR illumination.

\begin{figure}[t]
\centering
\includegraphics[width=1\textwidth]{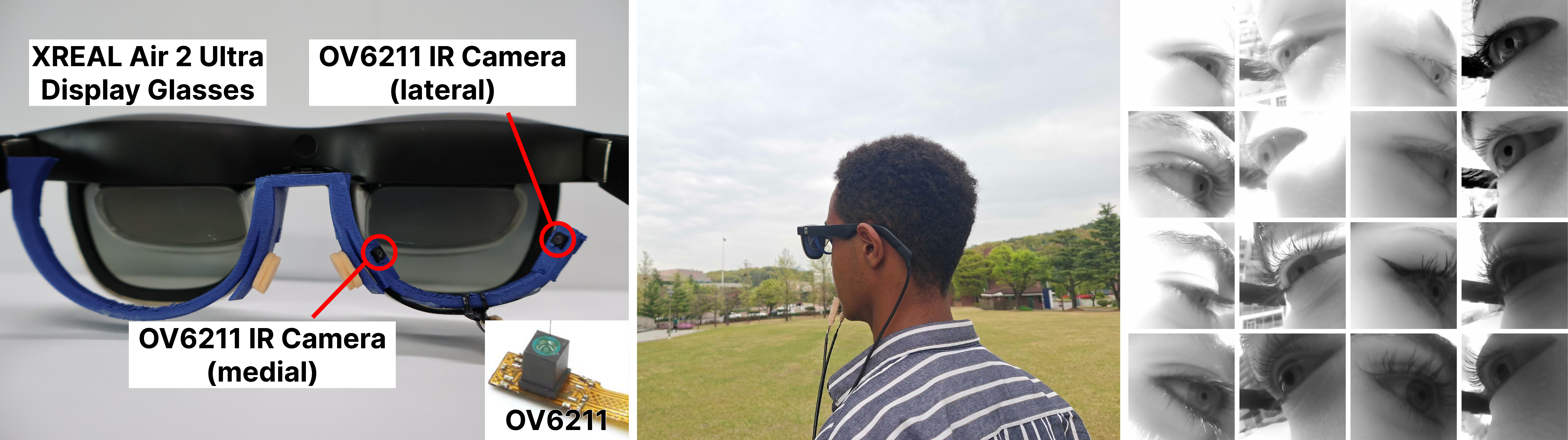}
\caption{Overview of the \datasetname data collection setup. Left: Custom eye-tracking glasses with OV6211 IR cameras mounted on 3D-printed holders at off-axis diagonal positions on the frame. Middle: A participant wearing the apparatus during outdoor data collection under the awaysun condition. Right: collected samples from the left column of the collected samples: lateral camera with facing sun, lateral camera with facing sun, medial camera with away sun, and medial camera with away sun.}
\label{fig:expsettings}
\end{figure}

\section{\datasetname Dataset}
As illustrated in Fig~\ref{fig:expsettings}, to account for different eye-camera placements in existing commercial eye trackers, we mount two IR cameras at distinct positions on the glasses, namely, lateral and medial. During the data collection, we cover diverse outdoor scenarios under varying lighting conditions. To ensure high-quality annotations, we first apply an object segmentation method and then manually check and correct each sample.

\subsection{Data Collection Device}
The experimental apparatus consisted of custom glasses built on the XREAL Air 2 Ultra, a commercial AR glass system\footnote{\url{https://www.xreal.com/}}, as a base platform. The built-in display in the glasses serves as our ground truth collection mechanism by presenting visual targets at known screen coordinates. Two OV6211 IR camera modules were mounted on the glasses frame via 3D-printed holders. We mounted two IR cameras for the right eye to simulate different mounting setups in existing commercial eyeglasses, including lateral (e.g., Pupil Neon~\cite{PupilNeon}, Tobii Glasses X~\cite{TobiiGlassesX}, Meta Aria~\cite{engel2023projectarianewtool}) and medial (e.g., Tonsen et al.~\cite{tonsen2020high}), as illustrated in Fig.~\ref{fig:expsettings}. This dual off-axis configuration enabled simultaneous capture of the right eye from two complementary viewpoints under identical illumination conditions.

The OV6211 camera modules captured monochrome images at $400\times400$ pixels resolution at 120 fps, incorporating an $850\pm10$ nm bandpass IR filter in a compact $6\times6\times3.5$ mm form factor, with 90-degree fields of view optimized for object distances of 20-50 mm. Critically, no dedicated IR illuminator is employed, so all eye imagery is captured exclusively under natural ambient illumination, distinguishing \datasetname from conventional datasets that rely on active IR light sources. The camera system is synchronized with the XREAL glasses' display system to ensure temporal alignment between captured eye images and ground truth target presentations on the display. We applied auto-exposure settings to both IR cameras.

\subsection{Collection Protocol}
We collected the data under variant lighting conditions and personal appearances, with accurate IR intensity measurement. 
Data were collected outdoors during daytime under two distinct natural lighting conditions designed to capture systematic variation in ambient near-infrared irradiance. Specifically, we defined a sun-exposed condition (\textit{sun-facing}), in which participants faced toward the sun, and a shadowed condition (\textit{sun-occluded}), in which participants faced away from the sun. Each participant completed both conditions sequentially in a fixed order, with sun-facing followed by sun-occluded. At the start of each recording, ambient near-infrared irradiance (mW/m²) in the 850--1000~nm band was measured using a High-Precision LED Phototherapy Light Meter from AquaHorti and recorded to characterize the lighting environment. Data were collected across ten days. Recording sessions took place outdoors during daytime hours, predominantly under clear, fair weather conditions.

We recruited 35 participants (19 male, 16 female; mean age 23.31 years, SD = 4.15) from a diverse set of backgrounds. The cohort spans 19 countries and includes participants from multiple ethnic groups, including Asian, White, Black or African, Middle Eastern or North African, Hispanic or Latino, and Eastern African backgrounds. This diversity allows the dataset to reflect a broad range of appearances and conditions encountered in real-world usage. Three participants wore eye makeup (e.g., eyeliner, mascara) during data collection, reflecting naturalistic variation in appearance in the dataset. 
During dataset collection, participants stood in their natural posture while holding an 8BitDo Micro controller\footnote{https://www.8bitdo.com/micro/} to register their responses, and there was no constraint on their head poses. A researcher stood beside each participant holding a laptop for the data recording.

For each sample, a white-filled circle appeared at a randomly selected location on the display and shrank continuously over one second from a radius of 80 pixels to 4 pixels. Participants were instructed to press any key on the controller when the circle reduced to a dot, after which a 500 ms inter-trial interval preceded the next trial. We recorded 80 samples for both \textit{sun-facing} and \textit{sun-occluded} conditions. Participants received 10\$ in compensation. This dataset collection was reviewed and approved by our institution, and all participants provided informed consent before participation.

\begin{figure}[t]
\centering
\includegraphics[width=1\textwidth]{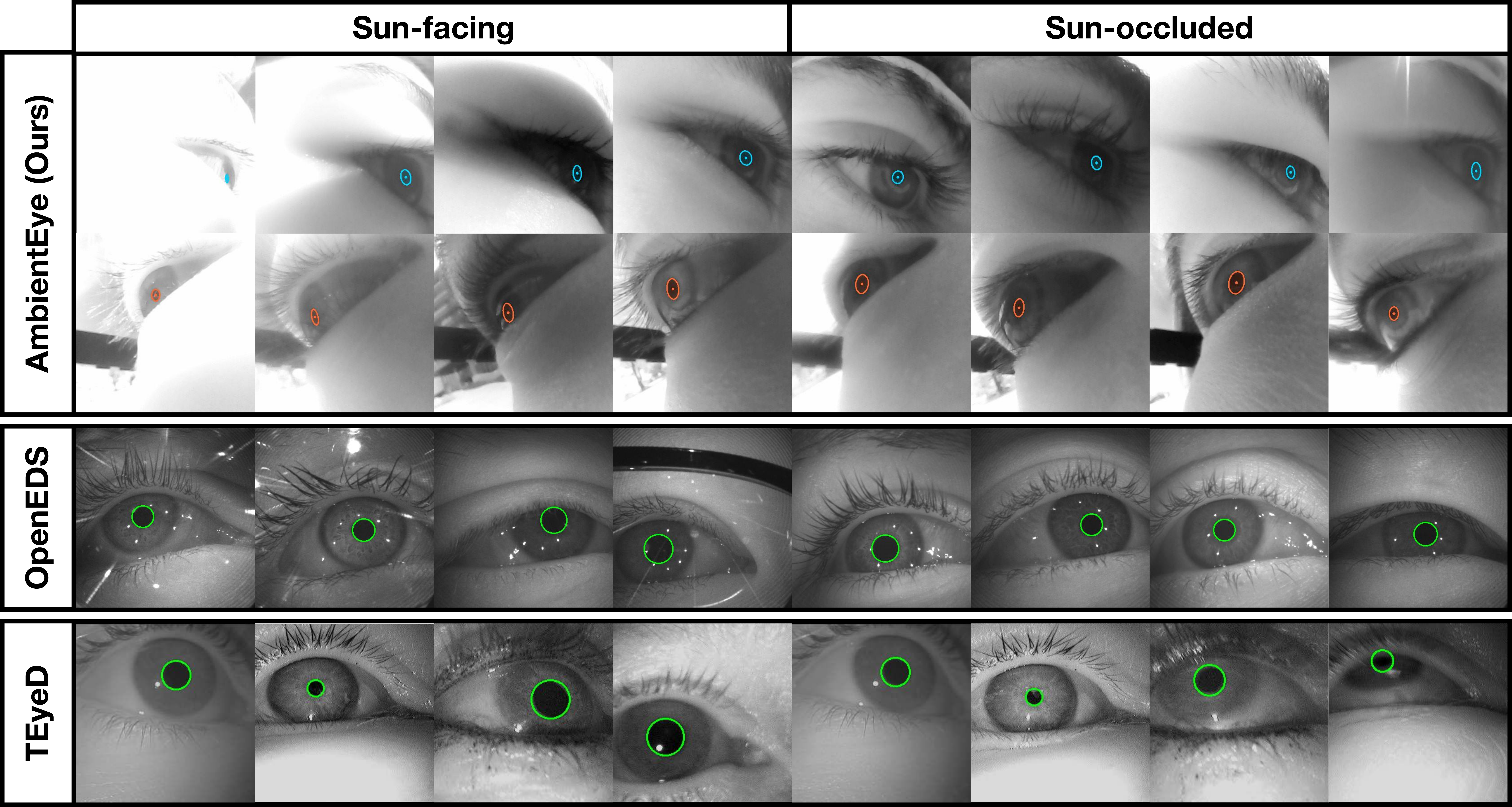}
\caption{Our \datasetname dataset captures diverse outdoor scenarios under sun-facing and sun-occluded lighting conditions from two camera viewpoints (\textcolor{cyan}{lateral}/\textcolor{orange}{medial}). The visualized circle represents the pupil contour, and the pupil center is denoted as a point computed from the center of the fitted ellipse. The first row shows samples from the medial view, and the second row shows samples from the lateral view. Samples from OpenEDS~\cite{garbin2019openeds} and TEyeD\cite{fuhl2021teyed} are presented along with their corresponding pupil annotation visualizations.}
\label{fig:data_sample}
\end{figure}

\subsection{Pupil Annotation}
Accurate pupil region segmentation or pupil center detection is critical for the gaze estimation task. To obtain high-quality pupil annotations, we adopt a two-stage process combining automated segmentation with human refinement. In the first stage, a single point is manually placed within the pupil region of the first frame of each session and used as a prompt for SAM2, as in prior work~\cite{Maquiling2025sampupil}, which then propagates the segmentation mask across all subsequent frames in the session. The resulting pupil regions are fitted with ellipses to provide an initial estimate. In the second stage, human annotators review and refine every frame of the segmentation result to validate its accuracy. When the predicted mask does not align well with the pupil boundary, annotators correct the annotation by manually marking pupil boundary points to fit an ellipse to the pupil. In total, pupil annotations were obtained for 2,518,693 out of 2,606,225 frames (96.6\%). Samples of the annotated data are shown in Fig~\ref{fig:data_sample}.

\section{Experiment}
The primary goal of \datasetname is to evaluate how well existing pupil segmentation methods perform under outdoor ambient-light conditions. The main motivation is that accurate pupil segmentation is the first and most critical step in most eye-tracking pipelines~\cite{liu2022eye}, and errors in pupil boundary localization directly limit downstream pupil center and gaze estimation accuracy. We therefore systematically benchmark the state-of-the-art segmentation model under \datasetname domain conditions.

\subsection{Datasets and Evaluation Protocol} 
We evaluate pupil segmentation using a state-of-the-art method, DenseElNet~\cite{kothari2021ellseg}. It is a representative model trained on six IR-based datasets, including OpenEDS~\cite{garbin2019openeds}, NVGaze~\cite{kim2019nvgaze}, RITEyes~\cite{nair2020rit}, LPW~\cite{tonsen2016lpw}, ExCuSe~\cite{fuhl2015excuse}, and PupilNet~\cite{fuhl2017pupilnet}, making it ideal to assess how well a model trained under controlled IR illumination adapts to the uncontrolled ambient IR conditions of our \datasetname. We use the intersection over union~(IoU) of the pupil as the evaluation metric following EllSeg~\cite{kothari2021ellseg}. We additionally evaluate EllSeg on three datasets: \datasetname (off-axis, ambient IR from sunlight), TEyeD~\cite{fuhl2021teyed} (on- and off-axis, active IR), and OpenEDS test set (on-axis, active IR). Since neither TEyeD nor \datasetname is included in the EllSeg training set, the evaluations correspond to within-dataset testing on OpenEDS, cross-dataset testing on TEyeD with active IR illumination, and cross-dataset testing on \datasetname without active IR illumination, reflecting progressively more challenging generalization tasks.

\subsection{Experimental Setup} 
Because the three evaluation datasets differ in native image resolution, we apply dataset-specific resizing to each before evaluation. Specifically, OpenEDS images are cropped around the scleral center from $640{\times}400$ to $400{\times}300$, then downsampled by a factor of $1.25$ to obtain the final $320{\times}240$ input. TEyeD images are downsampled by a factor of $1.2$ from $384{\times}288$ to obtain the final $320{\times}240$ input, and our images are zero-padded horizontally from $400{\times}400$ to $533{\times}400$ to match the $4{:}3$ aspect ratio, then downsampled to obtain the final $320{\times}240$ input. For \datasetname, we create a stratified sample of 2,376 frames per group across participants, matching the OpenEDS test set size and ensuring a balanced and fair cross-dataset evaluation. Experiments were implemented on a machine with an NVIDIA GeForce RTX 4090 GPU with 24GB VRAM.

\begin{figure}[t]
\centering
\includegraphics[width=1\textwidth]{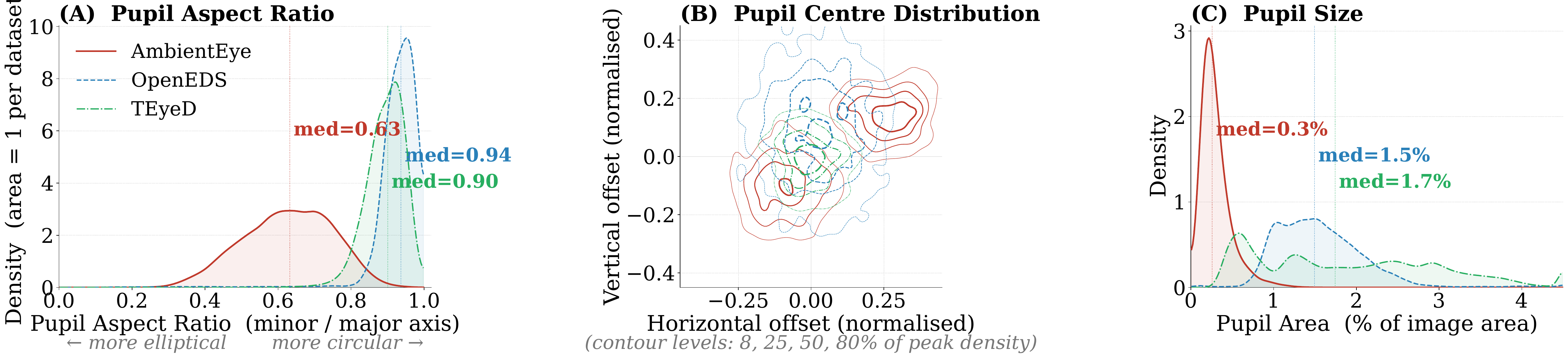}
\caption{(A) Probability density of the pupil aspect ratio (minor/major axis), computed over all annotated frames from each dataset. (B) 2D density of the normalized pupil center position relative to the image center. (C) Probability density of the normalized pupil area (ellipse area as a percentage of image area).}
\label{fig:pupil_ellipse_distribution}
\end{figure}

\begin{figure}[h]
\centering
\includegraphics[width=1\textwidth]{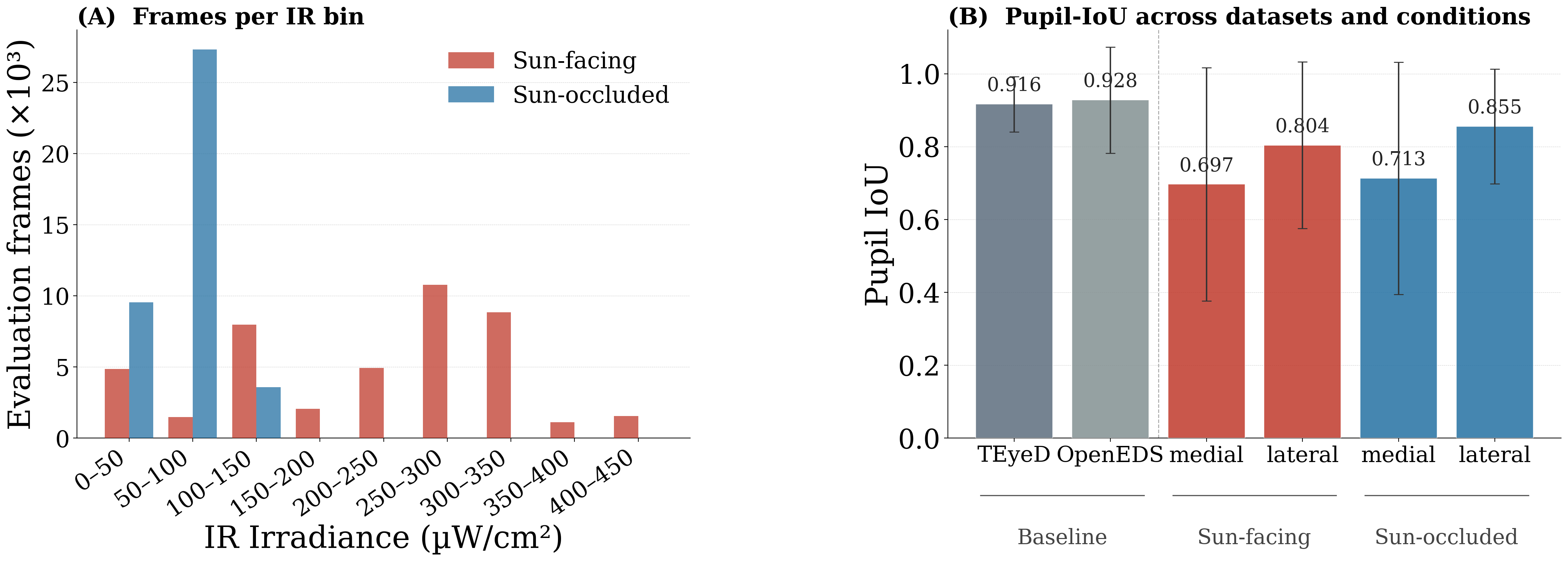}
\caption{IR irradiance analysis. (A) Frame count per IR intensity bin. (B) Pupil IoU across datasets and our conditions (sun-facing/sun-occluded, medial/lateral).}
\label{fig:segmentation}
\end{figure}

\begin{figure}[h]
\centering
\includegraphics[width=1\textwidth]{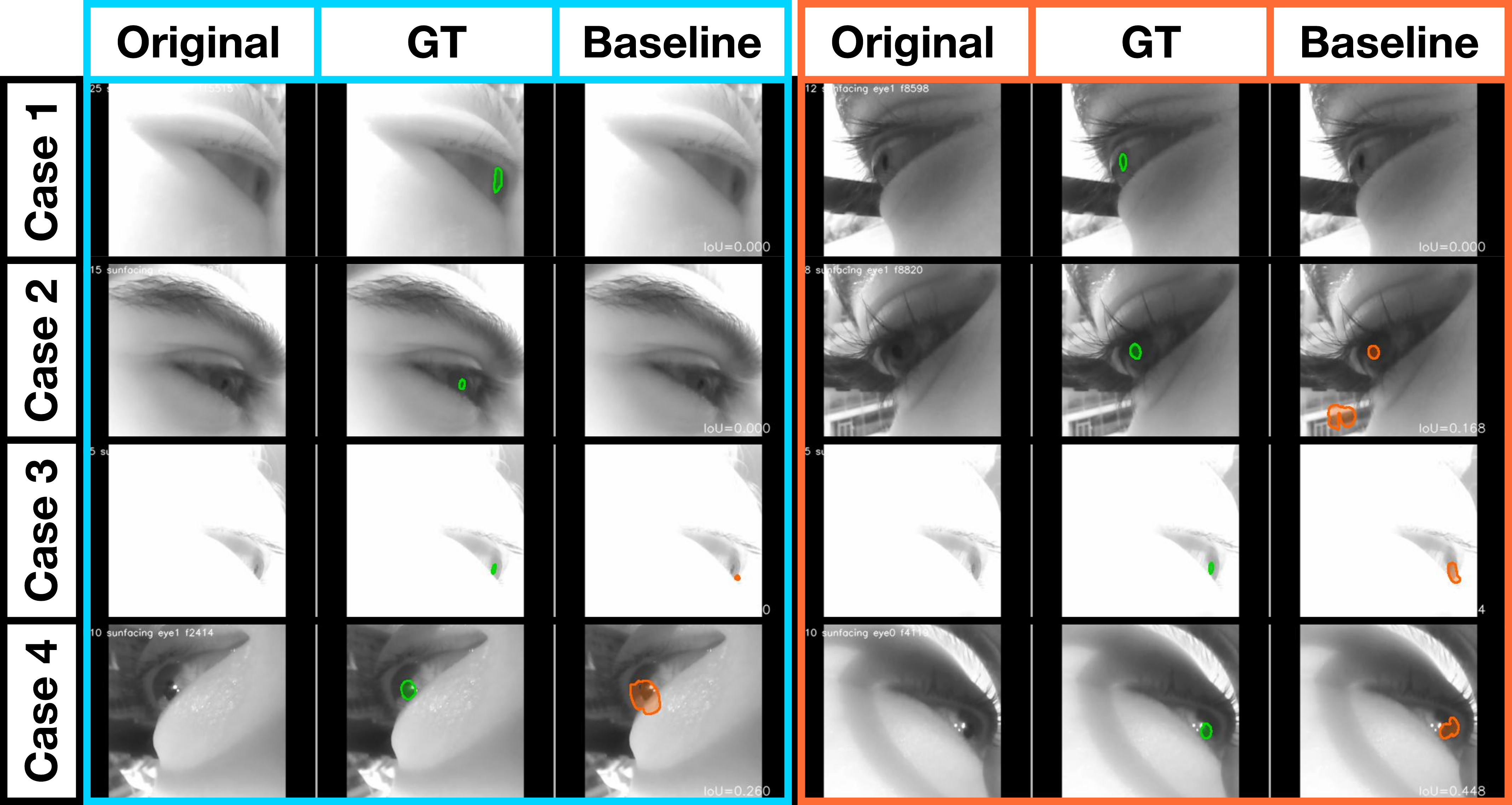}
\caption{Representative success and failure cases~(C\#) for each two axis~(\textcolor{cyan}{lateral} and \textcolor{orange}{medial}) in \datasetname. From top: original IR frame (left), GT contour in green (center), EllSeg prediction in orange (right) as a baseline condition. \textbf{C1}: highly off-axis ellipse distortion, lateral camera, aspect ratio\,$<$\,0.45. \textbf{C2}: intermediate aspect ratio (0.55--0.74) where failures occur without eye-closing. \textbf{C3}: high periocular brightness ($>$180\,px), sun-facing. \textbf{C4}: low solar altitude ($15\degree$--$30\degree$), grazing-angle IR maximises specular reflection.}
\label{fig:qualitative_cases}
\end{figure}

\subsection{Pupil Segmentation Analysis of Dataset}
As shown in Fig.~\ref{fig:pupil_ellipse_distribution}, three geometric characteristics distinguish \datasetname from existing datasets. First, pupils in \datasetname are substantially more elliptical (median aspect ratio\,=\,0.63) than those in OpenEDS (0.94) and TEyeD (0.90), reflecting the prevalence of near-circular, on-axis pupils in EllSeg's training data. Second, the pupil center distribution in \datasetname is far wider, reflecting greater camera off-axis variation inherent to wearable form factors. Third, the apparent pupil area is smaller on average (0.26\% of image area) than in OpenEDS (1.49\%) and TEyeD (1.74\%), reducing the number of pixels available for segmentation. Together, these three factors of extreme ellipticity, wide gaze-angle coverage, and small apparent size, highlight why \datasetname will present new challenges for models trained under controlled IR lighting conditions. 

\subsection{Pupil Segmentation Evaluation and Error Analysis}
\textbf{Overall Performance.}
As shown in Fig.~\ref{fig:segmentation}, EllSeg achieves strong performance on OpenEDS (IoU\,=\,0.928, within-dataset) and TEyeD (IoU\,=\,0.916, cross-dataset), confirming that the model generalizes well across controlled IR settings. However, performance drops substantially on \datasetname (overall IoU\,=\,0.767), reflecting the domain gap introduced by uncontrolled ambient illumination. Note that sun-facing frames are more challenging (medial\,=\,0.697, lateral\,=\,0.804) than sun-occluded frames (medial\,=\,0.713, lateral\,=\,0.855), consistent with the higher ambient IR irradiance under direct solar exposure.

\begin{figure}[h]
\centering
\includegraphics[width=1\textwidth]{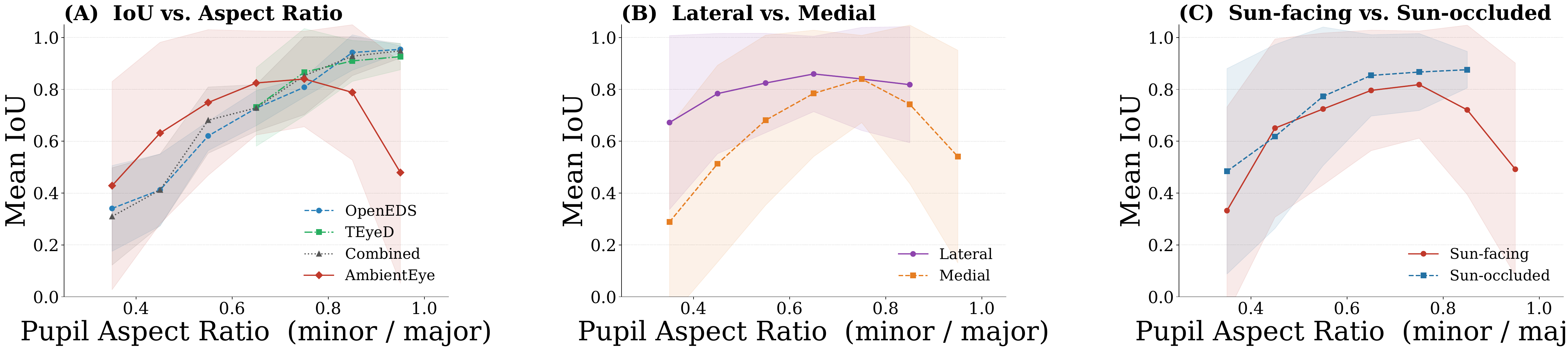}
\caption{IoU as a function of pupil aspect ratio across all evaluation sets. (A) OpenEDS, TEyeD, and \datasetname; shaded bands show $\pm1$ std. (B) \datasetname by camera: lateral vs.\ medial. (C) \datasetname by sun condition: sun-facing vs.\ sun-occluded.}
\label{fig:iou_by_aspect_ratio}
\end{figure}
\begin{figure}[h]
\centering
\includegraphics[width=1\textwidth]{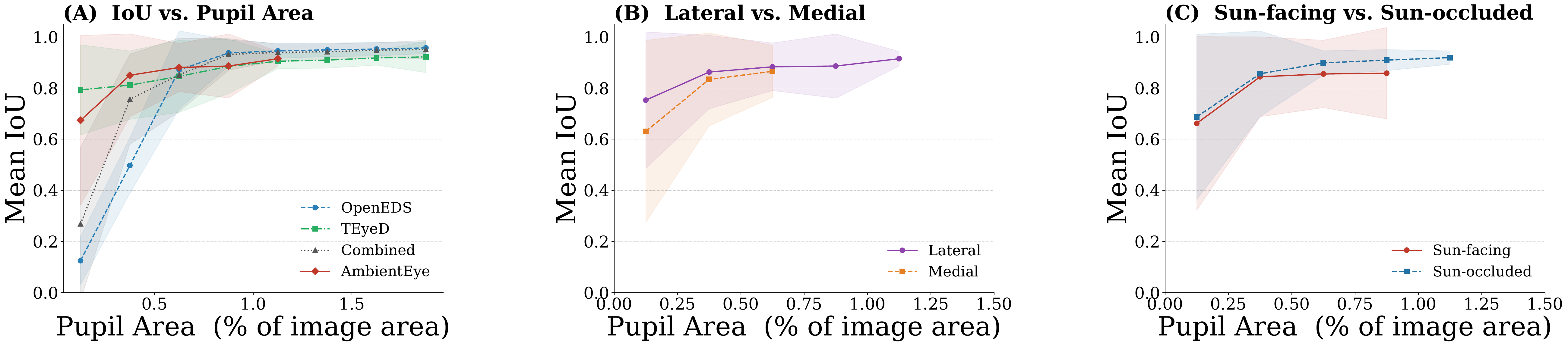}
\caption{IoU as a function of normalized pupil area (ellipse area as a percentage of image area). (A) OpenEDS, TEyeD, and \datasetname. (B) \datasetname by camera: lateral vs.\ medial. (C) \datasetname by sun condition: sun-facing and sun-occluded.}\label{fig:iou_by_area}
\end{figure}

\textbf{Pupil Size and Shape Analysis.}
We analyze how EllSeg IoU varies with pupil aspect ratio and area to identify the geometric drivers of the accuracy gap.
As shown in Fig.~\ref{fig:iou_by_aspect_ratio}A, both OpenEDS exhibit a pronounced IoU drop at low aspect ratios ($<$0.6). Frames in this regime are reported as eye-closing moments~\cite{fuhl2021teyed}, where the pupil appears as a thin horizontal slit and the model produces a degenerate prediction. The same behavior is visible in Fig.~\ref{fig:iou_by_area}A at very small pupil areas ($<$0.25\%), confirming that these two metrics co-vary for eye-closing frames. In \datasetname, the low-aspect-ratio and low-area regime contains both successes and failures, revealing a fundamentally different failure mode. Frames that succeed have a pupil that, while elliptical, is still visible with sufficient contrast; frames that fail show highly distorted shapes arising from extreme off-axis viewing geometry, the distribution documented in Fig.~\ref{fig:pupil_ellipse_distribution}A, which EllSeg has not encountered during training (Fig.~\ref{fig:qualitative_cases}, \textbf{C1}). Notably, failures persist even at intermediate aspect ratios (0.55--0.74) where eye-closing cannot account for the breakdown, indicating that off-axis distortion alone is sufficient to substantially impact model performance (Fig.~\ref{fig:qualitative_cases}, \textbf{C2}). The lateral camera consistently yields lower IoU than the medial at matched bins (Figs.~\ref{fig:iou_by_aspect_ratio}B and \ref{fig:iou_by_area}B), consistent with its larger off-axis angle. Sun-facing frames show lower IoU across both metrics (panel C), reflecting additional pupil constriction due to stronger ambient irradiance.

\textbf{Ambient Illumination Analysis.}
To quantify the effect of ambient illumination independently of pupil geometry, we evaluate EllSeg on 84,021 annotated frames sampled at one per 30 across all sessions. For each frame, we measure two brightness proxies: the mean number of grayscale pixels inside the ground-truth (GT) pupil contour on the original frame (\emph{GT pupil brightness}), capturing actual irradiance at the annotated pupil, and the mean number of grayscale of pixels inside the EllSeg-predicted pupil region at $320{\times}240$ (\emph{predicted pupil brightness}), capturing how the model perceives the pupil in its input space.

\begin{figure}[h]
\centering
\includegraphics[width=1\textwidth]{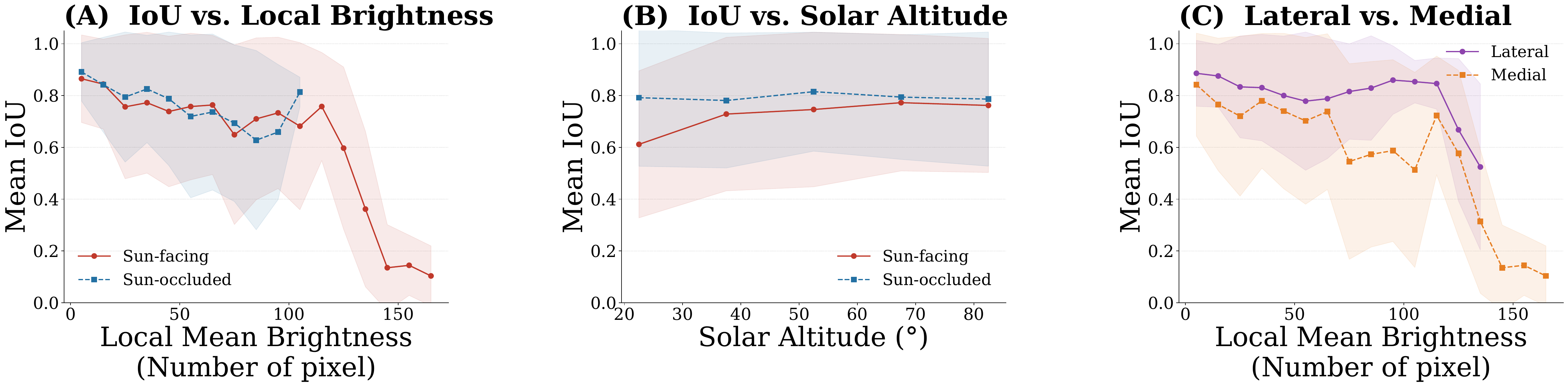}
\caption{IoU as a function of ambient illumination, evaluated on 84,021 annotated frames sampled at one per 30. (A) IoU vs.\ GT pupil brightness (mean grayscale inside the GT contour) for sun-facing and sun-occluded; shaded bands show $\pm1$ std. (B) IoU vs.\ solar altitude, binned at $15\degree$ intervals. (C) IoU vs.\ GT pupil brightness for lateral (eye1) vs.\ medial (eye0) cameras.}
\label{fig:iou_by_brightness}
\end{figure}

As shown in Fig.~\ref{fig:iou_by_brightness}A, IoU decreases monotonically with GT pupil brightness for sun-facing frames (0.865 at $<$10\,px to 0.104 at $>$140\,px), confirming that ambient IR saturation of the pupil progressively degrades segmentation. The collapse is sharp: IoU falls to 0.362 at 130--140\,px and to 0.135 at 140--150\,px, indicating a photometric threshold beyond which the dark pupil boundary is effectively erased. Sun-occluded frames are concentrated in the low-to-moderate range ($<$100\,px) with stable IoU (0.628--0.891), and rarely reach brightness levels sufficient to trigger saturation failure. Sun-facing frames extend to much higher GT brightness (up to 170\,px) because direct solar IR irradiates the pupil more intensely, and at every matched brightness bin, they exhibit greater variability than sun-occluded frames.

In the Fig.~\ref{fig:iou_by_brightness}B, the strongest impairment occurs at $20\degree$--$30\degree$, where sun-facing IoU falls to 0.612 versus 0.792 for sun-occluded. This grazing-angle regime maximizes specular reflection on the cornea and periocular skin, producing localized highlights that further corrupt the pupil boundary (Fig.~\ref{fig:qualitative_cases}, C4)
Fig.~\ref{fig:iou_by_brightness}C shows IoU as a function of GT pupil brightness split by camera. Lateral frames consistently outperform medial at matched brightness bins (0.886 vs.\ 0.842 at $<$10\,px; 0.833 vs.\ 0.720 at 20--30\,px). As GT pupil brightness increases, both cameras degrade, but medial shows a sharper collapse: IoU falls to 0.314 at 130--140\,px, 0.135 at 140--150\,px, and near-zero above 150\,px. The lateral camera rarely reaches this saturation regime (n\,=\,43 at 130--140\,px), as it is angled toward the nose, keeping the field of view confined to the face; the medial camera, in contrast, captures regions extending beyond the face into the background, and these out-of-face regions are occasionally misdetected as the pupil (Fig.~\ref{fig:qualitative_cases}, \textbf{C2} lateral). 

\textbf{Summary.}
Together, Figs.~\ref{fig:iou_by_aspect_ratio}--\ref{fig:iou_by_brightness}, with representative cases shown in Fig.~\ref{fig:qualitative_cases}, identify two distinct failure modes that are absent from controlled IR benchmarks. The first is \emph{geometric breakdown}: extreme off-axis viewing produces ellipse distortions outside the training distribution, causing failures both at low aspect ratios driven by lateral and medial camera position (\textbf{C1}) and at intermediate aspect ratios where the shape is no longer slit-like yet still challenges the model (\textbf{C2}). The second is \emph{photometric collapse} from ambient IR: high periocular brightness washes out the pupil--iris contrast (\textbf{C3}) and grazing-angle solar illumination at low altitudes introduces specular highlights that corrupt the boundary (\textbf{C4}). Their combined presence explains the large, structured performance gap in Fig.~\ref{fig:segmentation} and defines the central obstacle to all-day outdoor eye tracking on smart glasses without active IR illuminators.

\section{Conclusion}
We introduced \datasetname, the first large-scale dataset for pupil segmentation under passive ambient IR illumination: 2,518,693 annotated eye images from 35 participants across 19 countries, captured outdoors under natural sunlight with two off-axis camera viewpoints and two sun-orientation conditions. Benchmarking EllSeg reveals a substantial performance gap from 0.928 on OpenEDS to 0.767 on \datasetname, attributable to two failure modes absent from controlled IR benchmarks: \emph{geometric breakdown} from off-axis ellipse distortion (Fig.~\ref{fig:qualitative_cases}, C1--C2) and \emph{photometric collapse} from ambient IR saturation (C3--C4). We benchmark a single segmentation model under a zero-shot setting, and data were collected at one site in April with stationary participants on the right eye only, leaving seasonal, geographic, and motion variation as future work.

\bibliographystyle{unsrt}
\bibliography{references}  

\newpage
\end{document}